\documentclass[11pt,a4paper]{article}

\usepackage[T1]{fontenc} \usepackage[utf8]{inputenc}
\usepackage{amsmath,amssymb} \usepackage{textcomp}
\usepackage{booktabs} \usepackage{array} \usepackage{geometry}
\usepackage{placeins} \usepackage{graphicx} \usepackage{xcolor}
\usepackage[numbers,sort&compress]{natbib}
\usepackage{url}
\usepackage[breaklinks]{hyperref}
\hypersetup{
  colorlinks=true,
  linkcolor=blue,
  citecolor=blue,
  urlcolor=blue,
  bookmarksnumbered=false,
  pdftitle = {Climbing the Ladder: Automating Data Pipelines and Reproducing Baselines in Low-Resource Settings},
  pdfauthor = {Donato Meoli and Eryk Wdowiak},
  pdfsubject = {low-resource neural machine translation},
  pdfkeywords = {language models, low-resource NMT, Sicilian language},
}
\title{Climbing the Ladder:\\
       Automating Data Pipelines and Reproducing Baselines\\
       in Low-Resource Settings}

\author{Donato Meoli\\
        Dipartimento di Informatica, Universit{\`a} di Pisa\\
        Largo B.~Pontecorvo 3, 56127 Pisa, Italy\\
        {\texttt{\href{mailto:donato.meoli@outlook.com}{donato.meoli@outlook.com}}} 
        \and
        Eryk Wdowiak\\
        Arba Sicula\\
        Brooklyn, NY, USA\\
        {\texttt{\href{mailto:eryk@wdowiak.me}{eryk@wdowiak.me}}}
}

\date{October 21, 2026}

\begin{document}
\maketitle



\begin{abstract}
The massively multilingual translation systems that now include Sicilian
learned it from noisy, non-standardised web text.
To obtain better translation quality, we redevelop
the small neural machine translation (NMT) system that
\citet{wdowiak2022loresnmt} customized for Sicilian.
We rebuild on maintained software the extraction of the parallel text from the PDFs of the
bilingual journal \emph{Arba Sicula}, whose sentences are aligned by a language-agnostic sentence
encoder with thresholds tuned against an existing hand alignment and then filtered for the typical errors of automatic alignment. Furthermore, the tokenization and the desinence-biased subword segmentation of the original system are ported to a Perl-free implementation, and all the
results are measured on a frozen test set drawn from the journal. We compare small Transformers
trained from scratch with the pretrained NLLB model adapted by low-rank fine-tuning, in both
directions and with Italian as a closely related bridge language, and we adapt to the pretrained
model the reverse-training recipe of the original system, i.e., a stage on forward-scored
back-translations of the web Sicilian followed by a stage on the curated literary text. The
pretrained model is by far the better of the two, and the staged training improves the translation into Sicilian more than the translation from it. On the test set of the original system, and in the tokenized space in which it was scored, the staged model with a larger adapter and more back-translations is above the reverse-training recipe in the directions that involve English and below it between Sicilian and Italian. Yet, a multilingual adapter degrades the directions on which it was not trained, which limits its use as a general multilingual model.
\end{abstract}



\section{Introduction}
Languages without an army and navy often lack a standard orthography.
Without specific spelling rules and conventions, written texts in these languages
often exhibit great variability (sometimes even within a local dialect),
which inhibits the ability of conventional metrics to assess model performance
\citep{aepli2023benchmark}.
But even languages like Sicilian -- where published dictionaries have presented
a consistent orthography for several hundred years
\citep{pasqualino1785,mortillaro1876,nicotra1883,traina1868} --
can exhibit extreme orthographic diversity which
(when incorporated into a training dataset)
prevents a model from learning how to generate consistent text
and often causes the model to
generate foreign words instead of native words.

During the 13\textsuperscript{th} century,
the Sicilian School of Poets at the court of Emperor Frederick~II
created the first literary standard in Italy.  Their poetry
inspired Dante to write in his native Florentine, which later
became today's Italian language.
Following in their footsteps, poets like
Antonio Veneziano (1543-1593),
Giovanni Meli (1740-1815),
Domenico Tempio (1750-1821),
Ignazio Buttitta (1899-1997) and
Salvatore Camilleri (1921-2021)
continued the Sicilian literary tradition,
writing in a fairly consistent manner.
But writing is different from speaking, so
Giuseppe Pitrè (1841-1916)
developed a popular tradition which
recorded the common people
in their own words and in the sound of their own words.

One can hear echoes of the popular tradition in the daily speech
of millions of people in Sicily, Calabria and Puglia,
as well as in diaspora communities abroad.
And one can find echoes of the literary tradition in books,
journals and pages of the internet.
Nonetheless, the publicly available Sicilian language text
often exhibits extreme orthographic diversity which
once prevented practitioners from developing
datasets of Sicilian-language parallel text and once
caused Sicilian to be absent from every major translation service.

Sicilian is now covered by massively multilingual models such as NLLB-200 \citep{nllb2022};
however, its training text was mined from non-standard web sources, and the model has in effect
learned a non-standard orthography (cf.\ section~\ref{sec:standardization}).
Hence, the problem for Sicilian has moved from the absence of a sufficiently large dataset
(on which to train a model) to the quality of available training data.

High-quality material exists, especially in the bilingual literary journal \emph{Arba Sicula}
(published since 1979), which \citet{wdowiak2022loresnmt} used to create
the first Sicilian neural machine translation (NMT) system.
His work demonstrated that it's possible to develop a fairly good translation model
in low-resource settings, but it was a labor-intensive task.
He aligned sentences by hand using a recipe tailored to the
low-resource, morphologically rich setting.

This work rebuilds that effort with maintained,
mostly CPU-only tooling and seeks to obtain better quality
with modern pretrained models.



That system \citep{wdowiak2021recipe,wdowiak2022loresnmt} followed the low-resource
recipe of \citet{sennrich2019loresnmt}, i.e., a small and heavily regularized Transformer
\citep{vaswani2017attention} with a small subword vocabulary \citep{sennrich2016subword},
and biased the segmentation toward the desinences of Sicilian grammar;
its later reverse-training strategy \citep{wdowiakreverse}
builds a pretrained model of Sicilian in stages from synthetic and web text
before the fine-tuning on the literary one.

Traditionally, parallel text was automatically aligned with dictionary-based tools like hunalign.
More recently, practitioners have begun using multilingual sentence encoders
like LaBSE \citep{feng2022labse} and LASER3 \citep{heffernan2022laser3},
which were used to mine WikiMatrix \citep{schwenk2021wikimatrix} and the NLLB's Sicilian data.
Vecalign \citep{thompson2019vecalign} combines such embeddings with a dynamic program,
and collections such as OPUS \citep{tiedemann2012} gather the resulting corpora.

Pretrained multilingual models can be adapted to a specific pair at a small cost by low-rank
adapters \citep{hu2021lora}, and allow one to exploit closely related high-resource languages as
a bridge \citep{johnson2017zeroshot,arivazhagan2019massively,fan2020beyond,zhang2020improving};
also, back-translation of monolingual text \citep{sennrich2015backtrans} is the standard way of
adding in-domain synthetic data.  Section~\ref{sec:background} explains how one can use
such methods to overcome the challenges of low-resource translation,
as originally described by \citet{koehnknowles2017}.



Here, we extract the parallel text of the journal automatically, with a LaBSE-based
aligner whose thresholds are tuned against the second author's hand-aligned text.
The aligner's output is then filtered to remove any remaining errors.
And during training, the automatically aligned text and the hand-aligned sheets
are deduplicated and used together, because
the latter cover only a fraction of the pairs of each issue.

Furthermore, we port to Python the tokenization and the desinence-biased
segmentation of the original system, and add a lemma source factor \citep{sennrich2016linguistic}
to the from-scratch baseline. Finally, we fine-tune NLLB-200 with LoRA in both directions and
with Italian as a bridge, and we replace the custom pretraining of the reverse-training recipe
with the pretraining of NLLB, keeping its second stage (forward-scored back-translations of the
web Sicilian) and its last one (the curated literary text). Since the published results of the
original system were measured on a different test set and in the space of its tokenizer, we
score our models on that test set and in that space as well (cf.\ Table~\ref{tab:sota}).

The structure of the paper is as follows. Section~\ref{sec:background} recalls the models on
which all our systems are based. Section~\ref{sec:pipeline} describes the extraction, the
alignment and the cleaning of the parallel text and the assembly of the datasets, while
section~\ref{sec:standardization} discusses the orthography of the Sicilian data and the reason
why a multilingual encoder can align it at all. Tokenization and subword segmentation are the
subject of section~\ref{sec:tokenization}, the models and the staged training of
section~\ref{sec:models}, and the evaluation protocol of section~\ref{sec:evaluation}; the
results are reported in section~\ref{sec:results}, and section~\ref{sec:conclusion} concludes.



\section{Background}
\label{sec:background}
We now recall the model on which all our systems are based. NMT treats translation as a
conditional language model: an encoder reads the source sentence $x=(x_1,\dots,x_n)$, and a
decoder factorizes the probability of the target sentence $y$ from left to right as
\[
  p_\theta(y\mid x)=\prod_{t} p_\theta(y_t\mid y_{<t},x) \; ,
\]
where the parameters $\theta$ are trained by maximizing the log-likelihood of a parallel corpus,
and the translation is obtained by beam search. Its standard parameterization is the Transformer
\citep{vaswani2017attention}, which replaces recurrence with the scaled dot-product attention
$\mathrm{Attention}(Q,K,V)=\mathrm{softmax}(QK^{\top}/\sqrt{d_k})V$ (where $Q$, $K$ and $V$ are
the query, key and value matrices, and $d_k$ is the dimension of the keys), computed in several
parallel heads.

With little data a deep neural network overfits,
so we implement the best practices
of \citet{sennrich2019loresnmt}
who recommend a smaller model with high dropout
\citep{srivastava2014dropout} and a small subword vocabulary. Byte-Pair Encoding (BPE)
\citep{sennrich2016subword} gives an open vocabulary and shares statistics across the forms of a
word, which is especially useful for a morphologically rich language such as Sicilian. Further
improvements usually come from deeper models, when the data allow them
\citep{micelibarone2017deep}, and from linguistic input features \citep{sennrich2016linguistic},
which we use in section~\ref{sec:models}.



\section{Data Pipeline}
\label{sec:pipeline}
\subsection{Extraction from PDFs}
We read the text layer of the \emph{Arba Sicula} PDFs with PyMuPDF, which avoids the pdflatex
round trip of the original pipeline. However, the quality of this layer depends on the issue.
In fact, the PDFs of the issues up to AS18, and of AS21, are scans of the paper copies whose OCR text
carries no~accent over the vowels; from AS19 (1998) onward they were produced from the original computer
files. This date also marks (most likely) an improvement of the Sicilian itself,
since 1998 is the year of Privitera's \emph{Beginner's Sicilian} \citep{privitera1998},
the first Sicilian grammar published after the one of \citet{pitre1875},
which was followed by the more comprehensive grammar of \citet{bonner2001}; hence,
the scanned issues are kept out of the training data (they are used only for the complete
extraction of the archive mentioned below). Some of the born-digital issues (e.g., AS22 and AS23)
contain spaces of zero width inside the words, which we detect from the glyph positions and
remove before the text is read.

Each page is assigned a language according to its share of Sicilian and English stop-words, and
pages with too few tokens are left out. Since the journal usually sets Sicilian and English on
facing pages, each Sicilian page is compared with the English page on either side, and the
pairs are chosen greedily by decreasing LaBSE similarity (cf.\ section~\ref{sec:alignment}),
each page being used at most once; this corrects the pairings that a fixed left/right rule gets
wrong wherever a page is mislabelled or the order of the two languages is swapped.
Pages are then segmented into sentences with \texttt{pysbd} (with the Italian rules for Sicilian),
after the running headers and page numbers have been removed; fragments with fewer than two words or
mostly in capitals are dropped, and the words hyphenated at the end of a line are rejoined.



\subsection{Sentence alignment}
\label{sec:alignment}
We embed each sentence with LaBSE \citep{feng2022labse}, a language-agnostic dual encoder, and
build the cosine-similarity matrix
\[
  S_{ij}=\frac{\langle u_i,v_j\rangle}{\lVert u_i\rVert\,\lVert v_j\rVert} \; ,
\]
where $u_i$ and $v_j$ are the embeddings of the $i$-th Sicilian and of the $j$-th English
sentence. For a pair of pages, the similarity is the larger of the mean row maximum and the mean
column maximum of $S$, and the pair is kept if it is at least $\tau_{\text{page}}$ (with at least
two sentences on each page). Consecutive spreads are then chained into blocks, so that a sentence
cut by a page break is mended when it has no closing punctuation and the next page starts in
lowercase. Within a block, the sentences are aligned by a monotonic dynamic program over $S$
with moves $1$--$1$, $1$--$2$ and $2$--$1$, whose gain is the mean cosine of the sentences
involved, and moves $1$--$0$ and $0$--$1$, which cost a penalty of $0.5$; the program is banded
to the facing page and its neighbors (and falls back to the unbanded one if the band admits no
path). An aligned span is kept if its mean cosine is at least $\tau_{\text{sent}}$. This is in
the spirit of Vecalign \citep{thompson2019vecalign}, but requires neither a bilingual dictionary
nor hunalign.

Both thresholds were chosen on a grid (0.50, 0.55, 0.58 and 0.62 for $\tau_{\text{page}}$, 0.35,
0.40, 0.45 and 0.50 for $\tau_{\text{sent}}$) against the hand alignment of the second author
for the issues AS41 and AS42, which gave $(\tau_{\text{page}},\tau_{\text{sent}})=(0.50,0.40)$ (its F1 of 0.58 is 0.01 below that of
$\tau_{\text{sent}}=0.35$, which keeps more pairs at a lower precision).
On that gold the aligner reproduces 715 pairs exactly (recall 0.54, precision 0.63); these
figures are somewhat pessimistic, since a pair that is correct but segmented differently from
the hand alignment counts as an error, and since the hand alignment favored
the pairs that were easier to align.  Each pair records the printed Sicilian and English
page and its LaBSE similarity, which allows one to check it against the printed issue.



\subsection{Cleaning}
\label{sec:cleaning}
A review of the aligned text 
showed that it's easy to identify the aligner's errors,
which makes LaBSE's alignments much easier to perfect than hunalign's.
Because LaBSE's errors are easy to identify, we filter them as follows.
First, the residue of the page layout is removed from the sentences
(the InDesign slug of AS43 and the ``Return to the TOC'' links).
Then, a pair is dropped if
(i) the longer side has more than 1.6 times the words of the shorter one (when it has at least five words),
(ii) it contains a table-of-contents leader,
(iii) it contains two or more characters typical of OCR noise, or
(iv) more than 25\% of its lowercase English words are not in an English dictionary.
Note that the bound in (i) was first 2.0; the pairs between 1.6 and 2.0 have a
mean LaBSE similarity of 0.59, against 0.73 for those below 1.6,
which in our view justifies the tighter bound. After cleaning and deduplication,
the whole digital archive of the journal gives 15{,}465 pairs.



\subsection{Datasets}
\label{sec:datasets}
Two training sets are used (Table~\ref{tab:data}).
Our first set, on which the from-scratch models and the first NLLB experiments were trained,
joins our first extraction of 44 issues
(14{,}245 pairs, before the cleaning of section~\ref{sec:cleaning}) with the
``Good-Sicilian-in-NLLB'' set curated by the second author, i.e., sentences mined with LASER3
\citep{heffernan2022laser3,nllb2022} that were rescored and filtered for quality and for Corsican
contamination. Its it--scn part is the WikiMatrix subset \citep{schwenk2021wikimatrix}, with a
small curated ``Good-Sicilian-from-WikiMatrix'' set.

Our second set, used by the staged training of section~\ref{sec:staged}, replaces the web text
with curated sources. These are (i) the hand-aligned sheets of the second author, i.e., Dieli's
translation of Pitr\`e's fables and 23 issues of the journal (scn--en), Cipolla's translations
of Meli (scn--it), and the trilingual exercises of the grammars of Bonner and Cipolla, (ii) our
alignment of all the born-digital issues (AS19 to AS47, except AS21), since the sheets cover only 10 to 20\% of the pairs of
an issue and the two sources are therefore complementary, and (iii) the parallel pages of the
Young Sicilian Manifesto, of the Napizia website and of the Napizia Magazine,
from which one text under copyright was excluded.
Furthermore, the sheets provide 14{,}360 monolingual Sicilian sentences
(Cipolla's 2021 textbook) and a validation set of 121 trilingual lines. Its Sicilian side is
normalized as in section~\ref{sec:tokenization}, and the exact repeats are removed.

A validation and a test set of 1{,}000 pairs each were drawn once at random from our first extraction of \emph{Arba Sicula}, which provides the literary standard that FLORES does not, among the pairs with 3 to 80 words per side and a length ratio between 1/3 and 3, and then frozen; hence, all the scn--en results of section~\ref{sec:results} are computed on the same test set, except those on the test set of the second author (Tables~\ref{tab:sota} and~\ref{tab:recipe}, and the lower half of Table~\ref{tab:ablation}). The validation set of 1{,}000 pairs is used with the web set and to choose the epoch of the runs of Table~\ref{tab:ablation}, whereas the 121 lines of the sheets are the test set of the second author (cf.\ section~\ref{sec:evaluation}), and are never used for model selection, since the staged runs of section~\ref{sec:staged} train for a fixed number of epochs.
Each training set is then guarded against it: a pair is removed if its Sicilian side is a test
sentence (in raw or normalized form) or, in the second set, contains a test sentence of six or more
words; the validation pairs are removed in the same way from the web set and from the runs of Table~\ref{tab:ablation}, and the 121 lines of the second author from the second set. This guard removed 794 pairs from the second set (778 because of the test set and 16 because of the 121 lines of the second author), some of them from the hand-aligned sheets, which cover the same issues as the test set; without it, the results of the staged training would have been inflated.


\begin{table}[t]
\caption{Datasets (number of sentence pairs). ``Web'' is the first training set of
section~\ref{sec:datasets}, ``curated'' the second one; the test set is shared.}
\label{tab:data}
\centering
\begin{tabular}{llrl}
\toprule
set & direction & pairs & sources \\
\midrule
web     & scn--en & 27{,}386 & Arba Sicula (11{,}538) + Good-Sicilian-in-NLLB (15{,}848) \\
        & it--scn & 9{,}389  & WikiMatrix \\
curated & scn--en & 30{,}980 & hand-aligned sheets, Arba Sicula, Napizia \\
        & scn--it & 10{,}427 & hand-aligned sheets, Napizia \\
        & scn (mono) & 14{,}360 & Cipolla's second textbook \\
test    & scn--en & 1{,}000 & Arba Sicula \\
        & it--scn & 1{,}000 & WikiMatrix \\
\bottomrule
\end{tabular}
\end{table}



\section{Standardization}
\label{sec:standardization}
Although Sicilian does not have a single official orthography,
a common literary standard has existed for several hundred years
and provides a rough consistency, which we can refine
because the majority of the Sicilian sentences
in the NLLB dataset were written in that standard.

Nonetheless,
approximately one-third of the Sicilian sentences
that the NLLB team collected were written in local dialect.
Some of the sentences were written in Corsican.
And very few of the Sicilian sentences in the NLLB dataset
were paired with a valid translation.
We suspect that these data quality issues were caused
by seed data which is inconsistent with the literary standard.



\subsection{Why Sicilian lives in the Romance neighbourhood}
\label{sec:dataquality}
A pretrained multilingual encoder needs no explicit Sicilian support to \emph{read} Sicilian.
The representations it learns are not a shared interlingua that maps every language into one
abstract, universal meaning space; instead they cluster languages by linguistic similarity
\citep{kudugunta2019investigating} and organise around surface-form cues \citep{verma2026script}.
For Sicilian and Italian, which are close on both counts, the two accounts agree: Sicilian lands
in the Italian--Romance region either way.

The surface form of literary Sicilian is close to Italian, so reading it works well.
Retrieval on the multi-parallel FLORES-200 devtest \citep{nllb2022} confirms it: LaBSE
retrieves the correct English translation of a Sicilian sentence in 99\% of the cases
(precision@1), one point below Italian (100\%), which it supports officially. Furthermore, if one
discounts English (the pivot of LaBSE, to which all languages are close), the nearest
translation of a Sicilian sentence is Italian in 49\% of the remaining cases, more than twice as often as
French (21\%), the next Romance language.

The seed-data failure, however, is in the \emph{opposite} direction -- retrieving
the Sicilian translation of an English sentence.
Romance languages share a lot of vocabulary and grammar, so when a model
retrieves a Sicilian, Italian or Corsican translation of an English sentence,
the translation that it selects will reflect the surface-form cues in its training data.
Sometimes the NLLB seed data's surface form reflects the literary language in
Sicilian dictionaries, but other times it reflects local pronunciations, which
(we suppose) is what caused their model to frequently retrieve Corsican-language sentences
when assembling the NLLB's Sicilian corpus.

Since multilingual models encode script and surface form more than deeper linguistic structure
\citep{verma2026script}, the two directions behave differently for the same reason: an
Italianate surface form retrieves the right English sentence,
whereas seed data that look like Corsican lead the model
to generate the wrong Sicilian.
Hence, the proximity to Italian cuts both
ways. It allows our LaBSE-based aligner (cf.\ section~\ref{sec:alignment}) to work on a language
that LaBSE does not officially support; however, it gives no protection against contamination by
neighboring languages when a corpus is assembled by transfer only.

How easily a text is aligned depends, in turn, on how standard it is. We computed, text by text,
the LaBSE cosine of the hand-aligned gold pairs of AS41 and AS42: standard literary prose
(Cipolla's \emph{Don Chisciotti}) has a mean cosine of 0.76, with 91\% of its pairs above
$\tau_{\text{sent}}=0.4$, while dialect poetry has a mean of 0.64, with 80\% of its pairs above it.
Hence, the non-standard material is both the larger source of noise in the collected data and
the harder to recover, and curating the data toward the literary standard should improve quality
on both counts. For poetry, we also tried merging short consecutive verses into
longer units before scoring. We simulated it on the poetry pairs of the gold: merging up to three
verses raises the mean cosine by about 0.11 and the share above the threshold by 15 points (from 0.65 and 84\%, pooled over all the poetry pairs, to 0.76 and
99\%), although part of this gain is mechanical, since the merged units are fewer, longer and
more distinctive. This corrective is not yet part of the production aligner.


For a translation model, we need sentences written in the Sicilian
literary language and we need translations of those sentences.
So we back-translated all of the Sicilian sentences into English and Italian,
then we scored the resulting pair on the task of translation into Sicilian.
Because a sentence written in standard Sicilian will score better
than a sentence written in local dialect, forward-scoring the back-translations
allows us to identify the sentences written in standard Sicilian.
This method yielded 549,486 back-translations to simulate English-to-Sicilian
translation and 689,840 to simulate Italian-to-Sicilian translation.



\section{Tokenization and Subword Segmentation}
\label{sec:tokenization}
Because the majority of the Sicilian sentences in the NLLB dataset were
written in a common literary standard, we can refine them for consistency.
We first normalize the text toward
the standard established by Cipolla and augmented by Wdowiak for MT clarity,
i.e., strict use of L on definite articles (\emph{lu/la/li}),
Italian-style H on \emph{aviri} verbs (\emph{haiu, hai, havi, avemu, aviti, hannu}),
CI for \c ci (\emph{\c ciuri$\to$ciuri})
and R retained before a clitic (\emph{farlu}).
Our normalizer applies the documented spelling rules
(e.g., \emph{cchi\`u$\to$chi\`u}, \emph{io/iu/eu$\to$jo},
\emph{non$\to$nun}) at a ``std'' level, and further uncontracts and ASCII-folds the text at a
``full'' level. Since the tokenizer of NLLB was trained on un-normalized web Sicilian, it is not
obvious that normalization helps the pretrained model; hence, it was measured on
scn$\rightarrow$en, with the same English references for the three variants. BLEU is 30.67 on raw
text, 30.79 with ``std'' and 30.61 with ``full'' (chrF 56.31, 56.49 and 56.36, respectively).
Thus, the light ``std'' normalization gives the best scores, although by only 0.1 BLEU, which is within the
noise of repeated runs, while the ``full'' one is no better than raw text, which is consistent with a pretrained tokenizer
that expects un-normalized input. We therefore adopt ``std'' for all the later experiments.

For the from-scratch models, the Napizia tokenizer was reimplemented in Python (its output
coincides with that of the original Perl code, except that it also normalizes dashes and ellipses): it ASCII-folds the accents and uncontracts the
Sicilian forms (e.g., \^o$\to$\emph{a lu}, d\^u$\to$\emph{di lu}), which reduces sparsity. Then,
the joint scn--en BPE \citep{sennrich2016subword} is biased toward the textbook desinences
catalogued in the resources that Wdowiak assembled to standardize Sicilian (Dieli's
\emph{Sicilian Vocabulary}, Cipolla's \emph{Mparamu lu sicilianu} \citep{cipolla2013}, Bonner's
grammar \citep{bonner2001} and the \emph{Chi\`u d\^a Palora} dictionary), which are appended to
the Sicilian side when the merges are learned. This keeps the morphological endings as units; for
instance, Wdowiak's sentence \emph{Carinisi are dogs!} is segmented as
\texttt{car@@ in@@ isi are dogs !}. In his experiments this reduced the vocabulary from 14{,}000
to 2{,}000 units and gave the largest improvement in BLEU. We call this combination of tokenizer
and biased segmentation ``lever B'' in the following.



\section{Models}
\label{sec:models}
\subsection{From-scratch baseline}
The from-scratch baseline is a small Transformer trained on CPU with Sockeye-3
\citep{hieber2017sockeye}: 3 layers, model size 256, 4 attention heads, feed-forward size 1024,
a vocabulary shared by the two languages and a joint BPE of 4000 merges, i.e., about 6.7M
parameters. Dropout is 0.4 on the embeddings and 0.2 elsewhere, with label smoothing 0.1, and the model is trained with Adam (learning rate $2\cdot 10^{-4}$, batches of 1024 words) for at most 30
epochs, with early stopping after 8 checkpoints without improvement. Pairs with a side longer than 100 subwords are left out of training. On top of it we add, in this
order, lever B and a linguistic input feature (``lever D''). The latter is a
lemma source factor \citep{sennrich2016linguistic} summed into the source embedding as
\[
  e_j=E_{\text{word}}(w_j)+E_{\text{lemma}}(\ell_j) \; ,
\]
where $w_j$ is the $j$-th source word and $\ell_j$ its lemma; the lemma vocabulary is small
(about 900 entries), and therefore generalizes across the inflected forms of a word.
The lemmas are those of the lemma table of the original system, which maps 16{,}377 inflected
forms (mostly of verbs) to 1{,}252 lemmas; a word missing from the table receives an unknown lemma.

\subsection{Pretrained model}
Our pretrained model is NLLB-200 \citep{nllb2022}, a massively multilingual encoder--decoder that
supports Sicilian (\texttt{scn\_Latn}) and selects the target language through the first token
forced on the decoder. We adapt its 1.3B variant with LoRA \citep{hu2021lora}: the pretrained
weights $W_0$ of the query, key, value and output projections of the attention are frozen, and a
low-rank update is learned, which gives $W=W_0+\frac{\alpha}{r}BA$ with $r{=}32$, $\alpha{=}64$
and dropout 0.05 on the adapter (about 18.9M trainable parameters, i.e., 1.4\% of the model).
As usual, the base model is kept in fp16 and the adapters in fp32; training uses a learning rate of
$2\cdot 10^{-4}$ and sentences of at most 128 tokens, and decoding uses beam search with 5 beams (4 for the synthetic sentences below).
One adapter is trained on both directions at once (scn$\to$en and en$\to$scn),
and one on four (scn$\leftrightarrow$en and it$\leftrightarrow$scn).
The former is trained for two epochs, the latter for one.
For the back-translation experiment of section~\ref{sec:results},
we translate into Sicilian with the bidirectional adapter the 7{,}592 English sentences
left unaligned by the first extraction, keep the 7{,}498 synthetic pairs whose length ratio is
between 0.5 and 2 and whose Sicilian side is new, and train on them
and on the web set a further adapter for scn$\to$en, for one epoch, on top of the bidirectional one (merged into the base model).

\subsection{Staged training}
\label{sec:staged}
To obtain a pre-trained model, we can either develop a custom one from scratch
or we can download one from a model hub.  In this subsection, we explore both.
To build a custom pretrained model, we trained one in stages.
First, we trained a bidirectional Italian--English model on ParaCrawl,
then we further trained the same model on Sicilian forward-translations of that text.
In the next stage, we trained on back-translations from the NLLB's Sicilian text.
And in the final stage, we fine-tuned on \emph{Arba Sicula}.
We can similarly further train the NLLB-200 in stages,
but because the model already provides a multilingual pretraining,
we only need to perform the last two stages,
which require data prepared for this task
(distinct from the back-translations
mentioned in section~\ref{sec:standardization}).
Starting from the monolingual Sicilian of NLLB (1{,}057{,}469 sentences), we
back-translated it into English and Italian with the Sockeye models,
and scored each synthetic pair by the negative log-probability that
its real side is a translation of the synthetic one (the lower, the better);
since his models were trained on the literary standard,
a standard Sicilian sentence usually obtains a lower score than a dialectal one.
We did the same with the English side of the NLLB data (translated into Sicilian),
with 941{,}391 Italian sentences of WikiMatrix (translated into Sicilian)
and with the monolingual Sicilian of his sheets.
Our second stage takes, for each of these
files, the 100{,}000 pairs with the
best score
(all of them for the two files of his sheets, which have fewer)
among those that pass a few filters:
no unknown token,
at least five words on each side,
a source that is neither equal to nor a near-copy of the target (similarity below 0.9),
no overlap with the test set, and
a Sicilian side that does not end its words in -o more often than in -u, and has at most one word in -o
(an inexpensive test against Italian passing as Sicilian,
which tends to obtain the
best scores).
Together, the six lists cover the directions
scn$\to$en, en$\to$scn, it$\to$scn and scn$\to$it, and 100{,}000 natural it--en pairs of
WikiMatrix are added in both directions; the adapter is trained on them for one epoch.
Finally, the third stage continues the same adapter on the curated set of section~\ref{sec:datasets}
in the four directions scn$\leftrightarrow$en and scn$\leftrightarrow$it, for two epochs.

Furthermore, we also experimented with not discarding the pairs with the worst scores,
which contain the dialectal, Italian and Corsican text that the second stage filters out.
In these experiments we spent them in a first, coarse stage, which spreads the output probabilities
across the dialects and leaves broad peaks roughly in the right places.
We then allowed the later stages to sharpen them.
This curriculum prepends to the two stages above one epoch on the
200{,}000 worst-scoring pairs of each file (all of them for the two smaller files)
(with at least three words on each side and no filter on the endings),
and is otherwise identical to them, so that the two runs can be compared.



\section{Evaluation}
\label{sec:evaluation}
We score the systems with sacreBLEU \citep{post2018sacrebleu}, i.e., with BLEU
\citep{papineni2002bleu} (tokenization \texttt{13a}) and chrF \citep{popovic2015chrf}, on the
frozen literary test set of section~\ref{sec:datasets}. We do not use the FLORES test set of NLLB
for scn--en, since its Sicilian follows the orthography introduced after 2017 and comes from a
different, non-literary domain, and is therefore not representative of the \emph{Arba Sicula}
material; FLORES-200 devtest is used only for it$\leftrightarrow$en, which our test sets do not
cover. Sockeye systems are scored in tokenized space, as the upstream system was, whereas NLLB systems are scored on raw text, with the Sicilian side normalized as in training (except for the out-of-the-box rows marked as raw); the two groups of rows of Table~\ref{tab:results} are thus not strictly
comparable (the floor obtained by copying the source is 5.18 BLEU in the tokenized space and 5.44 on raw text). Training was
done on an 8-core CPU for the Sockeye models and on a single Colab GPU for the NLLB adapters. We
report no running times, since only the quality of the translation is of concern here.

For the comparison with the published results of the second author (cf.\
Table~\ref{tab:sota}), the NLLB systems are also scored on his test set in the six directions
between Sicilian, English and Italian. This set consists of the 121 trilingual lines that he
selected by hand from the issues AS38 and AS39 of the journal (labeled as validation in his
sheets). As he
did, we compute BLEU in the space of the Napizia tokenizer of the original system, i.e., on the
tokens that it produces from the hypothesis and from the reference (accents folded and
contractions undone, with the Italian side going through the English tokenizer). This space is
more lenient into Sicilian, since an accent or an apostrophe placed differently no longer costs a
match: on our test set it raises the en$\rightarrow$scn score of the staged model of
section~\ref{sec:staged} from 21.72 to 25.69 BLEU, and its scn$\rightarrow$en score only from
32.37 to 33.03. Since some of his lines also occur in our training data (8 in that of the staged
runs and 14 in the web set), we report the scores both on all 121 lines and on the lines absent
from the training data of each model. The re-decoding of a model reproduces its BLEU only up to a
few hundredths (e.g., 32.37 against 32.40 on scn$\rightarrow$en), since beam search in fp16 on
the GPU is not exactly deterministic.



\section{Results}
\label{sec:results}
Table~\ref{tab:results} reports the results on the frozen test set of 1{,}000 literary pairs. In
the table, ``floor'' is the score of the untranslated source, each row beginning with ``+'' adds
one component to the row above it, ``raw'' means that NLLB receives the Sicilian text without
normalization, and a dash marks a direction on which the model was not trained or evaluated. Note that the Sockeye baseline exceeds the floor by only 3.3 BLEU
(8.51 against 5.18, both in the tokenized space), that lever B raises it to 10.94, and that the lemma factor of lever D lowers it to 10.18. NLLB-1.3B, on the
other hand, is much better even out of the box (29.00 on scn$\to$en), and the bidirectional LoRA adapter trained on the web set raises it to 31.27, while en$\to$scn improves from 9.90 to 19.19. Adding the 7{,}498 back-translated
pairs gives 31.60 on scn$\to$en; this small improvement is consistent with the size of the
in-domain monolingual pool, which is two orders of magnitude smaller than the back-translations
used by the staged training.


\begin{table}[t]
\caption{Results on our test set (BLEU / chrF). The upper rows are trained on the web set, the
staged rows on the back-translations and on the curated set (cf.\ section~\ref{sec:staged}).}
\label{tab:results}
\centering
\begin{tabular}{lcc}
\toprule
model & scn$\rightarrow$en & en$\rightarrow$scn \\
\midrule
floor (copy source) & 5.18 / 25.95 & -- \\
Sockeye baseline & 8.51 / 32.73 & -- \\
\;+ lever B (tok.\ + desinences) & 10.94 / 34.66 & -- \\
\;+ lever D (lemma source factors) & 10.18 / 34.15 & -- \\
\midrule
NLLB-200 1.3B, out of the box (raw) & 29.00 / 55.23 & 9.90 / 41.09 \\
LoRA, bidirectional & 31.27 / 56.92 & 19.19 / 50.02 \\
\;+ back-translation (7{,}498 pairs) & 31.60 / 57.21 & -- \\
\midrule
staged (back-translations, then curated) & 32.40 / 58.19 & 21.72 / 52.79 \\
curriculum (coarse stage first) & 33.01 / 58.62 & 21.78 / 52.94 \\
\bottomrule
\end{tabular}
\end{table}


Staged training improves both directions, with 32.40 BLEU on scn$\to$en
and 21.72 on en$\to$scn; the gain over the bidirectional adapter is 1.13 BLEU on scn$\to$en
and 2.53 on en$\to$scn, which may be due to the back-translations having been selected for the quality of their Sicilian, i.e., of the target side of en$\to$scn. We remark that this
comparison mixes two changes, since the staged run replaces the web set with the curated one as
well as adding the back-translations; we have not trained the curated set alone.
Prepending the coarse stage of section~\ref{sec:staged} raises scn$\to$en to 33.01 BLEU (58.62 chrF), 0.61 more than the
staged run, while en$\to$scn is essentially unchanged (21.78 against 21.72). Repeated runs of the
bidirectional adapter differ by up to about 0.2 BLEU on scn$\to$en and 0.5 on en$\to$scn, and
therefore the first difference is probably real and the second is within the noise (however, the repetition of the staged run in Table~\ref{tab:ablation} differs from it by 0.42 BLEU on scn$\to$en, which makes the first difference less certain). Of course,
the curriculum also trains for one more epoch on about 826{,}000 pairs, and hence we cannot tell
whether its benefit comes from the dialectal text itself or from the longer training.



\paragraph{Ablation of the staged training.}
In order to find which settings limit the staged model, we have run it again with the settings of
section~\ref{sec:staged} (base) and with three changes made together. These are (i) 300{,}000
pairs per file in the second stage instead of 100{,}000, (ii) an adapter of rank 64 instead of 32 (with $\alpha{=}128$)
that also covers the feed-forward layers, and (iii) sequences of up to 256 tokens instead of 128
(at 128 tokens, about 2\% of the curated pairs are truncated), also in decoding, with three epochs in the third stage
instead of two. All runs hold out of training our test and validation sets and all the lines of
the test set of the second author, in the three languages. The epoch of the third stage is chosen
on the 1{,}000 validation pairs, and only that epoch is scored on the two test sets
(Table~\ref{tab:ablation}). The base run reproduces the staged run of Table~\ref{tab:results}
within 0.42 BLEU (32.82 and 21.52 BLEU against 32.40 and 21.72), and the three changes raise it to
35.40 BLEU on scn$\to$en and 25.09 on en$\to$scn, i.e., by 2.6 and 3.6 BLEU. On the test set of
the second author, the gain is 3.9 BLEU into Sicilian, 3.8 from it and 3.5 from Italian into
Sicilian, while scn$\to$it gains only 0.6. The two directions between English and Italian lose
1.8 and 2.2 BLEU instead; they are not trained in the third stage, and the longer stage possibly
makes them forget more. The larger second stage alone gives 33.34 and 21.88 BLEU on our test,
i.e., 0.5 and 0.4 more than the base run, which is about the noise of repeated runs; on the test
set of the second author it gains 1.4 BLEU from Sicilian into English and 2.7 from Italian into
Sicilian, and changes the other directions by about 1 BLEU at most. Adding the larger adapter gives
34.60 and 24.18 BLEU on our test, i.e., 1.3 and 2.3 more than the larger second stage alone, and
the longer sequences with the third epoch add a further 0.8 and 0.9; hence, most of the gain comes
from the adapter. On the test set of the second author, however, the variant without the longer
third stage is better in five directions out of six (e.g., 62.02 against 60.95 BLEU from Sicilian
into English, and 61.89 against 59.09 from Italian into English), and it is worse only from English
into Sicilian (47.64 against 48.36). Since the variants are compared on the validation set, which
favors the longer third stage (33.17 and 23.28 BLEU against 32.54 and 22.07), we keep the latter in
Table~\ref{tab:sota}. Finally, since it$\to$en and en$\to$it are not trained in the third stage,
we have also added to it 10{,}427 natural WikiMatrix it--en pairs per direction that the second
stage did not use (last column), reusing the same second stage. On our test the scores do not
change (35.49 and 25.16 BLEU), and on the test of the second author scn$\to$it gains 0.9 BLEU
(62.94), but it$\to$en and en$\to$it do not recover (57.30 and 47.53 against 59.09 and 48.00),
possibly because the encyclopedic text of WikiMatrix is far from the literary one of his test.
Since its validation score is slightly lower (33.19 and 22.89 BLEU), this variant is not kept. Also, the validation score was still rising at the third epoch
(23.28 BLEU into Sicilian against 22.06 at the second), which suggests that a longer third stage
may improve the model further. The final model therefore keeps the three changes, adds to the third
stage the 9{,}389 WikiMatrix it--scn training pairs of the web set in both directions (whose
validation and test pairs stay out) and allows up to four epochs, reusing the same second stage
(column ``final''). Its validation score still rises at the fourth epoch (33.58 and 23.83 BLEU),
the highest of all the variants, and on our test it reaches 36.01 BLEU on scn$\to$en and 26.34
on en$\to$scn, i.e., 0.6 and 1.3 more than the three changes. On the test of the second author,
however, it gains only from English into Sicilian and into Italian (49.25 against 48.36, and 48.77
against 48.00), and it loses 0.7 to 1.8 BLEU in the other directions, among which the two between
Sicilian and Italian that the WikiMatrix pairs were meant to help (61.19 and 60.30 against 62.78
and 62.05). Since it is the best variant on the validation set, it is the model of
Table~\ref{tab:sota}.


\begin{table}[t]
\caption{Ablation of the staged training (cf.\ section~\ref{sec:staged}): the base settings,
the larger second stage alone, the larger second stage with the rank-64 adapter, three changes made together (300{,}000 back-translated pairs per file, a rank-64 adapter
on the attention and feed-forward layers, 256-token sequences and three epochs in the third
stage), the latter with 10{,}427 natural it--en pairs per direction in the third stage, and the final model, i.e., the three changes with the WikiMatrix it--scn pairs in the third stage and up to four epochs. The epoch is chosen on the validation set; our test is scored on raw text (BLEU / chrF),
his on all 121 lines in his tokenized space (BLEU), since all runs exclude them from training.}
\label{tab:ablation}
\centering
\footnotesize
\setlength{\tabcolsep}{3pt}
\begin{tabular}{lcccccc}
\toprule
 & base & + 300k pairs & + rank 64 & + three changes & + it--en & final \\
\midrule
epoch chosen & 2 of 2 & 2 of 2 & 2 of 2 & 3 of 3 & 3 of 3 & 4 of 4 \\
\midrule
\multicolumn{7}{l}{our test} \\
scn$\rightarrow$en & 32.82 / 58.41 & 33.34 / 58.75 & 34.60 / 59.89 & 35.40 / 60.39 & 35.49 / 60.41 & 36.01 / 60.70 \\
en$\rightarrow$scn & 21.52 / 52.68 & 21.88 / 52.97 & 24.18 / 54.70 & 25.09 / 55.24 & 25.16 / 55.09 & 26.34 / 55.75 \\
\midrule
\multicolumn{7}{l}{test of the second author} \\
en$\rightarrow$scn & 44.42 & 44.49 & 47.64 & 48.36 & 48.46 & 49.25 \\
scn$\rightarrow$en & 57.14 & 58.54 & 62.02 & 60.95 & 59.81 & 59.34 \\
it$\rightarrow$scn & 59.31 & 62.04 & 64.42 & 62.78 & 61.86 & 61.19 \\
scn$\rightarrow$it & 61.46 & 61.02 & 62.67 & 62.05 & 62.94 & 60.30 \\
it$\rightarrow$en  & 60.90 & 59.97 & 61.89 & 59.09 & 57.30 & 58.42 \\
en$\rightarrow$it  & 50.16 & 49.14 & 48.76 & 48.00 & 47.53 & 48.77 \\
\bottomrule
\end{tabular}
\end{table}



\paragraph{Italian directions.}
Italian is closely related to Sicilian and has much more data, and NLLB supports
it$\leftrightarrow$scn natively. On the web set, we hold out a frozen it--scn test set of 1{,}000
pairs from the WikiMatrix data, evaluate NLLB-1.3B out of the box, and then fine-tune the four
directions (scn$\leftrightarrow$en and it$\leftrightarrow$scn) jointly into one multilingual
adapter, which uses Italian as a bridge in the sense of \citet{johnson2017zeroshot} and
\citet{fan2020beyond}. Table~\ref{tab:italian} reports the results as BLEU / chrF. As one would
expect for a close, high-resource neighbor on in-domain WikiMatrix text, the Italian pairs are by
far the ``easiest'': scn$\rightarrow$it reaches 43.47 BLEU, and it$\rightarrow$scn improves from
17.57 to 26.97. However, the multilingual adapter loses somewhat on scn$\leftrightarrow$en (30.75
and 17.49 BLEU, against 31.27 and 19.19 for the bidirectional adapter of
Table~\ref{tab:results}); hence, on the web set the Italian bridge brings no improvement to the
English directions.


\begin{table}[t]
\caption{Trilingual adapter on the web set, out of the box and after fine-tuning (BLEU / chrF). The
scn--en rows use our test set, the it--scn rows the WikiMatrix one, and the it--en rows the
FLORES-200 devtest, on which the adapter was not trained.}
\label{tab:italian}
\centering
\begin{tabular}{lcc}
\toprule
direction & out of the box & multilingual FT \\
\midrule
scn$\rightarrow$en & 29.00 / 55.23 & 30.75 / 56.57 \\
en$\rightarrow$scn & 9.90 / 41.09 & 17.49 / 48.78 \\
it$\rightarrow$scn & 17.57 / 44.76 & 26.97 / 51.69 \\
scn$\rightarrow$it & 42.23 / 59.05 & 43.47 / 59.79 \\
it$\rightarrow$en & 34.42 / 62.65 & 27.14 / 57.92 \\
en$\rightarrow$it & 29.42 / 58.63 & 28.77 / 58.07 \\
\bottomrule
\end{tabular}
\end{table}


\paragraph{Forgetting of the untrained directions.}
Since the adapter is trained only on the directions that involve Sicilian, one may ask what happens to the others. On FLORES-200, the trilingual adapter lowers it$\to$en from 34.42 to 27.14
BLEU and en$\to$it from 29.42 to 28.77 (Table~\ref{tab:italian}), i.e., it forgets a direction
on which the base model was good and on which it was never trained. This is the transfer from
high- to low-resource languages described by \citet{johnson2017zeroshot}, and it limits the use
of the adapter as a general multilingual model; the translation between Sicilian and the other
languages of NLLB (e.g., Spanish, French or Portuguese) is probably affected as well. On the
FLORES-200 devtest, the trilingual adapter gives 7.8 to 11.4 BLEU into Sicilian from these three
languages and 19.1 to 30.1 BLEU from Sicilian into them; however, we have not scored the base model
on these directions, and hence we cannot tell how much of it is lost.



\paragraph{Relation to the state of the art.}
\label{sec:sota}
Table~\ref{tab:sota} compares our NLLB systems with the published results of the second author
on his test set and in the tokenized space in which he scored them (cf.\
section~\ref{sec:evaluation}). Its first three columns are his, i.e., the baseline and the larger
many-to-many (m2m) model of \citet{wdowiak2022loresnmt} (the latter trained with back-translations
and, for the Italian directions, with the trilingual textbook exercises), and the third stage of
his reverse-training system \citep{wdowiakreverse}. The other columns are ours, i.e., the
bidirectional adapter trained on the web set (the adapter of Table~\ref{tab:results}, whose re-decoding scores 31.30 and 19.19 BLEU on our test set against 31.27 and 19.19) and the two staged runs of
section~\ref{sec:staged}, each scored on all 121 lines and on those absent from its training
data. On the latter, the staged model is on a par with the reverse-training system from English
and from Italian into Sicilian (45.07 against 45.1 BLEU, and 61.49 against 61.4), and it exceeds
it by 7.8 BLEU from Sicilian into English and by 2.0 and 14.3 BLEU between English and Italian;
it remains below it only from Sicilian into Italian (59.84 against 62.9). The baseline of 2022 (25.1
on en$\rightarrow$scn and 29.1 on scn$\rightarrow$en) is exceeded by about 20 and 27 BLEU, and
already the web adapter exceeds it by 6.9 and 21.6 BLEU. The curriculum is somewhat better than the
staged run on scn$\rightarrow$en (58.19 against 56.43) and somewhat worse on
en$\rightarrow$scn (43.92 against 45.07), which on 113 lines is probably within the noise. The
final model of the ablation (cf.\ Table~\ref{tab:ablation}) raises the score into Sicilian
to 49.25 BLEU and the one from it to 59.34, i.e., 4.2 and 10.7 BLEU above the reverse-training
system; it remains below it from Italian into Sicilian (61.19 against 61.4) and from Sicilian into
Italian (60.30 against 62.9), and it loses somewhat between English and Italian w.r.t.\ the staged run. The
gap of Table~\ref{tab:results} between our en$\rightarrow$scn score and his baseline (21.72
against 25.1) may therefore be due to the test set and to the metric rather than to the model: on
raw text, the staged model obtains 39.43 BLEU into Sicilian on the clean lines of his test set
and 21.72 on ours, and the tokenized space adds a further 3 to 6 BLEU into Sicilian (cf.\
section~\ref{sec:evaluation}).

It is fair to remark, however, that the comparison is not entirely even. The baseline of the
second author chose its checkpoints on this same test set (training stopped when the score on it
ceased to improve), and so did his later systems, whereas our staged runs train for
a fixed number of epochs and were scored on these lines only after training. On the other hand,
his training data do not contain the test lines (although they contain other text of the same articles), whereas ours do. Also, his test file exists in several revisions, which are only intermediate stages of its tokenization.
Finally, the first stage of his recipe, i.e., the adaptation of an Italian--English model to
Sicilian, is the part that we have not reproduced; on his test set its role seems to be taken by
the multilingual pretraining of NLLB, with the possible exception of scn$\rightarrow$it, the only
direction in which his system remains ahead.


\begin{table}[t]
\caption{Published results of the second author against ours on his test set, in his tokenized
space (BLEU). His columns are taken from \citet{wdowiak2022loresnmt} and \citet{wdowiakreverse};
in ours, the first number is the score on all 121 lines and the second the score on the lines
absent from the training data of that model (107 for the web adapter, 113 for the staged runs);
the last column is the final model of Table~\ref{tab:ablation}, which was trained without any
of the 121 lines and is therefore scored on all of them. The web adapter was trained on scn$\leftrightarrow$en only, and a dash marks a direction that was
not reported.}
\label{tab:sota}
\centering
\small
\setlength{\tabcolsep}{4pt}
\begin{tabular}{lccccccc}
\toprule
 & \multicolumn{3}{c}{Wdowiak} & \multicolumn{4}{c}{ours (NLLB-200 1.3B)} \\
\cmidrule(lr){2-4} \cmidrule(lr){5-8}
direction & baseline & m2m & reverse & web & staged & curriculum & final \\
\midrule
en$\rightarrow$scn & 25.1 & 35.0 & 45.1 & 32.19 / 31.95 & 44.40 / 45.07 & 43.97 / 43.92 & 49.25 \\
scn$\rightarrow$en & 29.1 & 36.8 & 48.6 & 51.16 / 50.65 & 57.00 / 56.43 & 58.80 / 58.19 & 59.34 \\
it$\rightarrow$scn & --   & 36.5 & 61.4 & 39.69 / 39.50 & 61.04 / 61.49 & 60.69 / 60.86 & 61.19 \\
scn$\rightarrow$it & --   & 30.9 & 62.9 & 52.98 / 52.57 & 60.40 / 59.84 & 60.47 / 59.96 & 60.30 \\
it$\rightarrow$en  & --   & --   & 48.2 & 56.59 / 55.63 & 62.72 / 62.45 & 61.38 / 61.11 & 58.42 \\
en$\rightarrow$it  & --   & --   & 46.7 & 47.31 / 45.89 & 49.27 / 48.69 & 49.14 / 48.62 & 48.77 \\
\bottomrule
\end{tabular}
\end{table}



\paragraph{Reproduction of the original baseline.}
In order to separate the effect of the model from that of the data, we have also retrained the
2022 baseline of the second author with his configuration, as recorded in his repository. This
uses Napizia tokenization, separate BPE vocabularies of 3{,}000 merges per language (the Sicilian one
learned with the inflections of the Dieli and Chiu da Palora vocabularies appended), and a
Transformer with 3 layers, model size 256, 4 heads and feed-forward size 1{,}024, whose target
embeddings are tied to the output layer. Dropout is 0.5 on the embeddings and 0.25 elsewhere,
with label smoothing 0.1, and the model is trained with Adam (learning rate $1.5\cdot 10^{-4}$,
batches of 20 sentences) for at most 20 epochs, with Sockeye 3 instead of his Sockeye 1. Pairs with a side longer than 200 subwords are left out, as in his configuration (his later many-to-many model used 100). We
train it on three sets, i.e., (A) his hand-aligned sheets, (B) our automatic alignment of the
issues from AS19 to AS47 with the textbook exercises, and (C) both, deduplicated, each guarded against
his test set and against the validation lines, and we score it on his test set in the tokenized
space, as in Table~\ref{tab:sota}; Table~\ref{tab:recipe} reports the results.

On his data, the reproduction remains well below the published scores (15.97 against 25.1
BLEU on en$\rightarrow$scn, and 16.65 against 29.1 on scn$\rightarrow$en). Two differences of
protocol may account for at least part of the gap: he started each model from a previously
trained one rather than from scratch, and he stopped the training on the test set,
whereas we stop on 500 lines of our validation set. Also, none of our six runs converged within
the 20 epochs (the validation score was still improving when the cap was reached), which suggests
that a longer training, such as the one implied by his warm start, would raise them. Our
alignment (B) gives 2.8 to 3.5 BLEU less than his sheets, which is consistent with its
smaller size and with the absence of Dieli's translations of Pitr\`e; however, the two together
(C) give 22.45 and 20.69 BLEU, i.e., 6.5 and 4.0 more than A, and therefore the
automatic alignment contributes pairs that the hand alignment does not contain. Of course, all
these scores remain far below those of the NLLB adapters on the same lines (Table~\ref{tab:sota}).


\begin{table}[t]
\caption{The 2022 baseline of the second author retrained from scratch with his configuration,
on three training sets, and scored on his test set in his tokenized space (BLEU). The first row
is the published score of \citet{wdowiak2022loresnmt}, obtained with a warm start and with the
training stopped on the test set.}
\label{tab:recipe}
\centering
\begin{tabular}{lrcc}
\toprule
training set & pairs & en$\rightarrow$scn & scn$\rightarrow$en \\
\midrule
published (his sheets) & -- & 25.1 & 29.1 \\
A (his sheets) & 20{,}220 & 15.97 & 16.65 \\
B (our alignment) & 18{,}503 & 13.20 & 13.14 \\
C (both) & 31{,}730 & 22.45 & 20.69 \\
\bottomrule
\end{tabular}
\end{table}



\section{Conclusion and Future Work}
\label{sec:conclusion}
We have rebuilt, on maintained software and without Perl, the path from the PDFs of \emph{Arba
Sicula} to a Sicilian--English NMT system, i.e., the extraction, alignment and cleaning of the
parallel text, the tokenization and desinence-biased segmentation of the original system with
an added lemma source factor, and the adaptation of NLLB-200 with LoRA, trained in stages on the
back-translations of the second author and on the curated literary text. On our test set the
best model, i.e., the staged training with three times the back-translations, a larger adapter,
longer sequences and the WikiMatrix it--scn pairs, reaches 36.01 BLEU on scn$\rightarrow$en and 26.34 on en$\rightarrow$scn,
while the coarse stage proposed by the second author helps the translation from Sicilian, but not
the one into it. On the test set of the second author and in his tokenized space, the same model
reaches 49.25 BLEU into Sicilian and 59.34 from it, i.e., 4.2 and 10.7 BLEU above his
reverse-training system, and it remains below it only between Sicilian and Italian. Hence, the
low score into Sicilian on our test set is probably due to the test set and to the metric more
than to the model. Retrained from scratch with his configuration, the original baseline does not reach its
published scores on his data, while our alignment added to his sheets raises it by 4.0 to 6.5
BLEU.

Several extensions are open. In particular, the aligner could merge the short verses of poetry,
as simulated in section~\ref{sec:dataquality}, and could collect the Sicilian left unaligned
as well as the English, which would enlarge the in-domain pool for back-translation.
Also, the training data could be weighted by quality,
e.g., favoring the issues published after 1998 or the texts written or
translated by Cipolla, whose style varies less than that of the whole journal (the hard part
being to identify them, since the issues carry no such metadata). If the back-translations of
the web Sicilian turn out to be of little use, forward-translations produced by a small model
trained on the literary text are an alternative for the pretraining stage.
Furthermore, the forgetting of the untrained directions could be limited
by mixing into the adaptation a large multi-parallel corpus
(e.g., Bible translations in many directions), at the price of a much longer training. It would
also be interesting to compare the encoder representations of translated sentences under
bidirectional and unidirectional training, and at different depths, 
and to measure how much the textbook exercises contribute.
Finally, the same pipeline may be applied to varieties close to Sicilian,
such as Calabrese and Salentino, whose written texts exhibit similar orthographic diversity.
The aligner itself should transfer to other varieties of Italy: on the FLORES-200 devtest, LaBSE
retrieves the correct English translation of a Friulian, Ligurian, Lombard, Sardinian or Venetian
sentence in 98.5 to 99.9\% of the cases (99.0\% for Sicilian), and therefore what is missing for
them is a bilingual source comparable to \emph{Arba Sicula}.

Our code for the pipeline and the evaluation is publicly available at
\url{https://github.com/dmeoli/SicilianNMT}, together with a small
serving application (a web interface and a Telegram bot, both built on a FastAPI service over
the NLLB adapter). The scored back-translations of section~\ref{sec:staged} are released on
Hugging Face, one dataset per direction (\url{https://huggingface.co/datasets/Napizia}); the text
of the journal is under copyright, and the hand-aligned sheets are not public, so the curated set
is not released.



\section*{Acknowledgments}
Arba Sicula, Gaetano Cipolla and Arthur Dieli developed the resources that
made machine translation possible for the Sicilian language.
We thank them for their support and encouragement.
And we thank them for allowing us to use
their digital collections for this project.



\pdfbookmark[0]{References}{sec:references}
\hypertarget{sec:references}{}
\bibliography{references}


\end{document}